\PassOptionsToPackage{hyphens}{url}
\documentclass[10pt,a4paper]{article}
\usepackage{zju-preprint}
\usepackage[numbers,sort&compress]{natbib}
\usepackage{xspace}
\usepackage{array,multirow,mathtools}
\usepackage{colortbl}
\usepackage{tabularx}
\usepackage{algorithm}
\usepackage{algpseudocode}
\newcommand{\method}{GameWeaver\xspace}

\newcommand{\up}{\ensuremath{\uparrow}}
\newcommand{\dn}{\ensuremath{\downarrow}}

\newcommand{\R}{\mathbb{R}}
\DeclareMathOperator*{\argmin}{arg\,min}

\renewcommand{\PaperTitle}{GameWeaver: Joint Video--Camera Diffusion for Automated Game World-Model Data Production}
\renewcommand{\PaperShortTitle}{GameWeaver}
\renewcommand{\PaperAuthors}{Lifan Jiang\quad Yifan Huang\quad Jixuan Zhang\quad Peifeng Wang\\[1pt]
  Yuying Ma\quad Dehai Zhao\quad Deng Cai\quad Boxi Wu\textsuperscript{\textdagger}}
\renewcommand{\PaperAffiliations}{Zhejiang University}
\renewcommand{\PaperContact}{wuboxi@zju.edu.cn}
\renewcommand{\PaperAuthorNote}{\textsuperscript{\textdagger}Corresponding author: Boxi Wu.}
\renewcommand{\PaperAbstract}{Recent advances in world models for game interaction and control have increased the demand for large-scale gameplay videos paired with camera trajectories. However, capturing synchronized camera trajectories often requires game-specific pipelines, limiting scalability. We therefore propose \method, a unified diffusion framework that automatically produces paired video--camera data through three complementary tasks: camera pose estimation from observed video, video generation along a prescribed camera trajectory, and joint video--camera generation from a reference image and text. First, each tipxel stores the normalized 3D coordinates of a pixel ray's endpoint in a reference coordinate frame. The resulting three-channel tipxel fields and RGB frames share the pretrained video model's VAE. Next, a fully trainable dual-stream Transformer exchanges appearance and geometry features through bidirectional gated attention and conditions both streams on shared text features. Finally, deterministic fitting recovers camera intrinsics and normalized trajectories from generated tipxel fields. In our evaluation tasks, \method achieves a normalized translation MSE of $0.0029$ for camera control; for joint generation, it scores $85.48$ out of 100 on instruction adherence versus $64.71$ for the best cascade on that metric, validating the feasibility of the proposed approach.
}
\hypersetup{pdfauthor={Lifan Jiang, Yifan Huang, Jixuan Zhang, Peifeng Wang, Yuying Ma, Dehai Zhao, Deng Cai, Boxi Wu}}

\begin{document}
\MakePreprintTitle
\begin{figure*}[t]
\centering
\includegraphics[width=\textwidth]{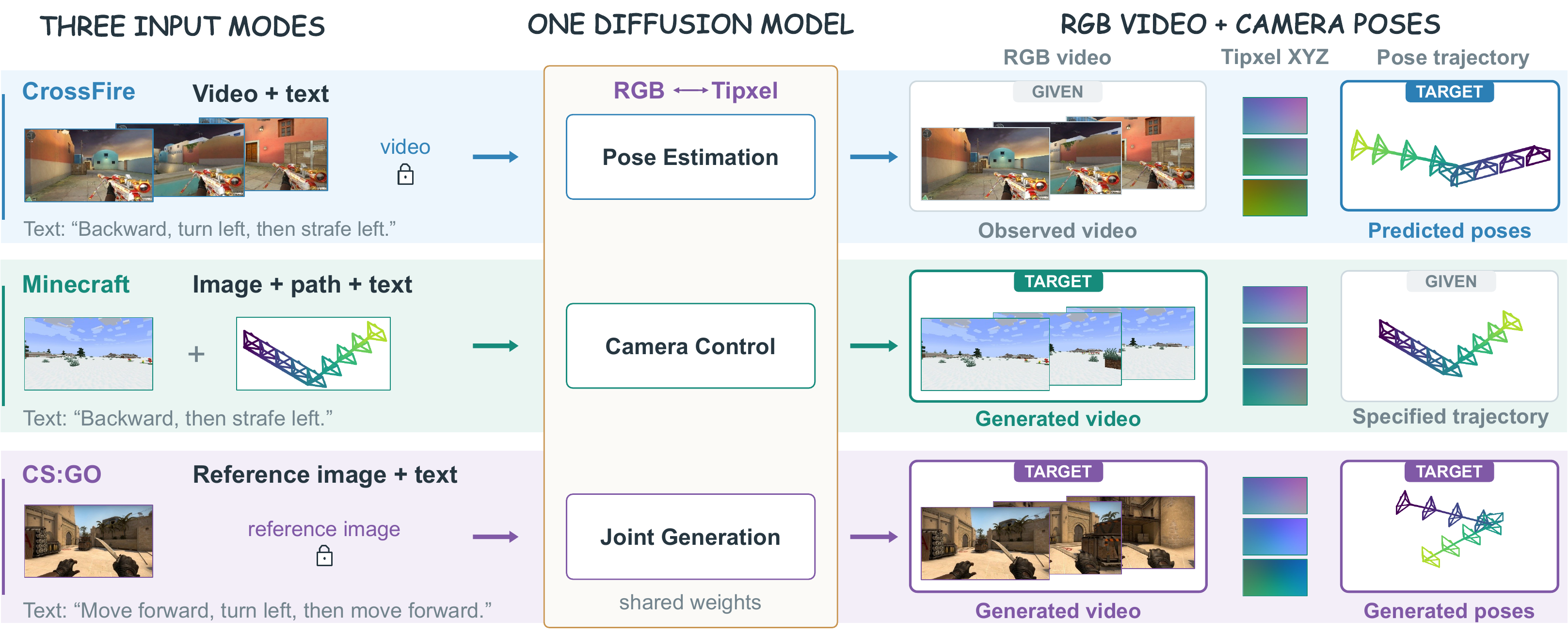}
\caption{\textbf{One diffusion model for three tasks.} Pose estimation predicts cameras from video and text. Camera control generates video from a reference image, a given trajectory, and text. Joint generation predicts video and cameras from a reference image and text. TARGET marks predicted outputs and GIVEN marks observed inputs. The resulting pairs support camera-conditioned world-model training.}
\label{fig:teaser}
\end{figure*}

\section{Introduction}
\begingroup
\setlength{\parfillskip}{0pt plus 0.45\linewidth}
Training camera-conditioned game world models~\cite{solarwm,lingbot,sanawm} requires large collections of gameplay videos paired with camera trajectories. Collecting accurate trajectories alongside the video typically requires a capture pipeline tailored to each game engine.

One way to produce paired data is to generate a video and recover its camera trajectory. Another is to plan the trajectory and generate a video conditioned on it~\cite{vipe,gendop,gen3c,scope}. In the first cascade, geometric distortions in generated frames can impair pose estimation. In the second, the motion in the generated video may deviate from the planned trajectory.

We propose \method, a flow-based model of visual appearance and camera geometry. Figure~\ref{fig:teaser} shows its three tasks: camera pose estimation from observed video, video generation along a prescribed camera trajectory, and joint video--camera generation from a reference image and text.

We represent each camera as a three-channel tipxel map. The pretrained video VAE encodes both these maps and the RGB frames. We construct a fully trainable dual-stream Transformer that exchanges appearance and geometry information through bidirectional gated attention, with shared text features conditioning both streams. Task-specific masks specify which latents stay fixed and which the model generates. Geometric fitting recovers camera intrinsics and normalized trajectories from the generated tipxel maps. Pairing the recovered cameras with observed videos annotates existing recordings, while pairing them with generated videos produces new training examples.

For pose estimation, \method has lower mean translation error than COLMAP within each game. For camera control, it achieves the lowest normalized translation error among the methods evaluated. For joint generation, it leads the compared cascades in motion-calibrated consistency, motion-normalized smoothness, and instruction adherence (Sec.~\ref{sec:experiments}). We also compare joint and factorized generation and examine the effects of cross-modal attention, noise schedules, and the velocity loss.

Our contributions to the production of paired gameplay data are threefold:
\begin{itemize}
\setlength{\emergencystretch}{3em}
\item \textbf{A unified data-production framework.} We introduce \method, which unifies pose estimation, camera control, and joint generation in one flow-based model for producing game world-model training data.
\item \textbf{A camera representation compatible with the video VAE.} The video VAE encodes three-channel tipxel maps. Fitting the decoded maps recovers camera intrinsics and a trajectory in the reference camera frame.
\item \textbf{A dual-stream architecture for appearance--\allowbreak geometry interaction.} We design a dual-stream Transformer with bidirectional gated attention and shared text conditioning to jointly model video and cameras. We evaluate all three tasks and compare gated cross-modal attention with attention over concatenated video and camera tokens.
\end{itemize}
\par\endgroup

\section{Related Work}
\begingroup
\setlength{\parfillskip}{0pt plus 0.45\linewidth}
\textbf{Game world models and training data.}
World Models learns latent dynamics for control~\cite{worldmodels}. DIAMOND and GameNGen build diffusion game simulators~\cite{diamond,gamengen}. Genie learns interactive environments through latent actions~\cite{genie}. SolarWM, LingBot-World, and SANA-WM develop camera-conditioned world models~\cite{solarwm,lingbot,sanawm}, increasing demand for paired supervision. Built on Wan2.2~\cite{wan,wan22}, \method supplies paired data by annotating gameplay recordings and synthesizing new clips.

\noindent\textbf{Camera estimation and control.}
COLMAP uses geometric reconstruction~\cite{colmap}; PoseDiffusion, RayDiffusion, and DiffusionSfM model cameras through diffusion~\cite{posediffusion,raydiffusion,diffusionsfm}. MASt3R, VGGT, DA3, and Pi3 infer geometry~\cite{mast3r,vggt,da3,pi3}; ViPE recovers video cameras~\cite{vipe}. CameraCtrl and MotionCtrl condition video on camera motion~\cite{cameractrl,motionctrl}; GEN3C, Uni3C, SCoPE, DualCamCtrl, and WorldStereo incorporate geometric controls~\cite{gen3c,uni3c,scope,dualcam,worldstereo,hyworld}. GenDoP generates camera trajectories~\cite{gendop}. \method couples appearance and geometry during denoising to perform both camera estimation and camera control.

\noindent\textbf{Joint multimodal generation.}
UniDiffuser unifies conditional and joint distributions through modality-specific noise levels~\cite{unidiffuser}. Diffusion Forcing supports flexible sequence generation with independent token noise~\cite{forcing}, and flow matching learns generative vector fields~\cite{flowmatching}. Rays as Pixels~\cite{jang2026rays} learns pose estimation and view synthesis from real-world videos with reconstructed cameras. It can jointly generate video and camera trajectories from images. \method produces training pairs for game world models from a reference image and an action prompt, using camera supervision recorded during rendering.
\par\endgroup

\section{Joint Video--Camera Diffusion}
\label{sec:method}
\begin{figure*}[t]
\centering
\includegraphics[width=\textwidth]{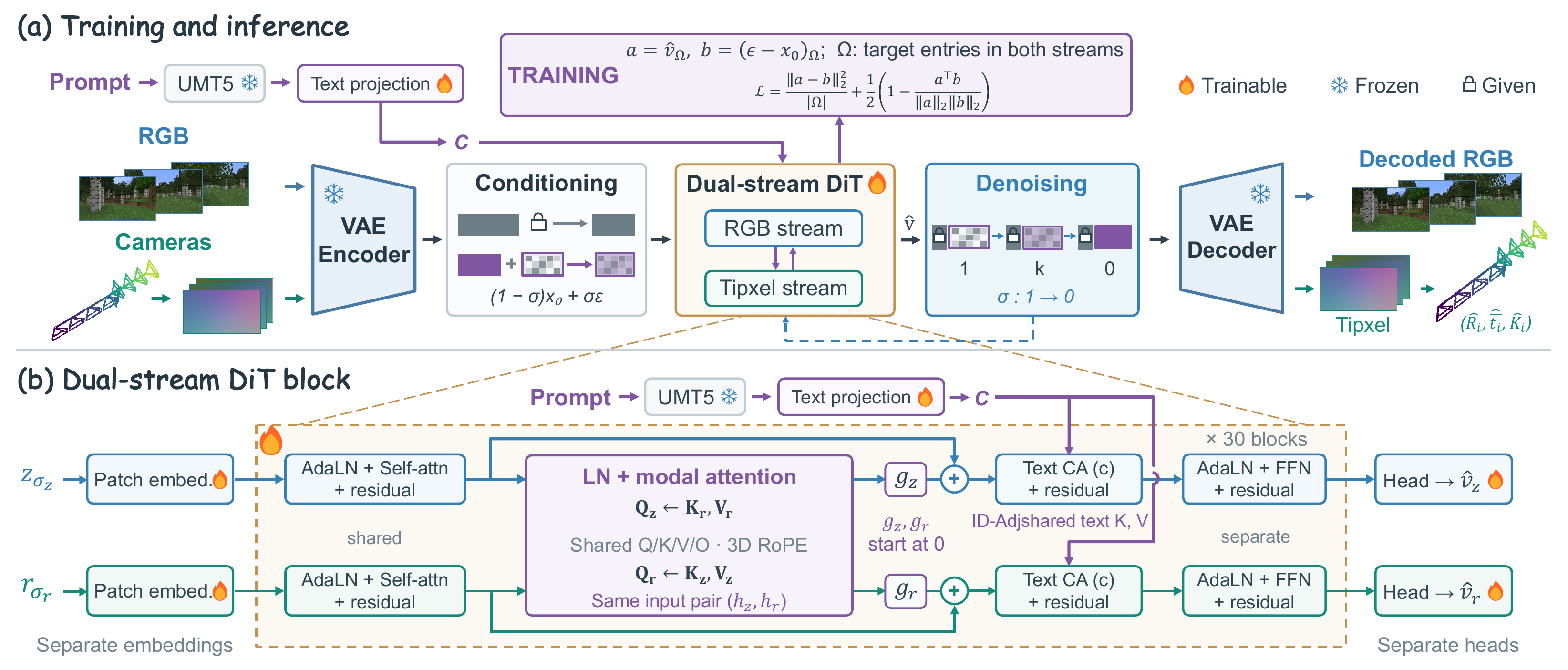}
\caption{\textbf{Training and inference with RGB and tipxel maps.} \textbf{(a)} Training fixes the known latents and interpolates noise separately in each stream. Velocity supervision covers the unknown elements $\Omega$, with $|\Omega|=M$ in Eq.~\ref{eq:loss}. At inference, the frozen VAE decodes sampled latents. Fitting the decoded tipxel maps yields rotations, normalized translations, and intrinsics. Pose estimation retains the observed RGB frames. \textbf{(b)} Bidirectional attention and zero-initialized residual gates couple the streams in each DiT block. Projected text $c$ supplies keys and values to both text cross-attention layers. The streams share self-attention and text-attention weights within each block, with separate feed-forward networks and output heads. Flames mark trainable weights, snowflakes frozen weights, and locks fixed conditions.}
\label{fig:architecture}
\end{figure*}

\subsection{Problem and data-production interfaces}
Let $V=(I_0,\ldots,I_{F-1})$ be a video, $p$ a text prompt, and $C=\{(R_i,t_i,K_i)\}_{i=0}^{F-1}$ its cameras. The rotation $R_i$ and camera center $t_i$ express camera $i$ in the source camera's coordinate frame, and $K_i$ is its intrinsic matrix. We learn a joint model over video and camera fields, with three conditional interfaces:
\begin{align}
\text{pose estimation:}\quad &P_\theta(C\mid V,p), \nonumber\\
\text{camera control:}\quad &P_\theta(V\mid I_0,C,p), \label{eq:interfaces}\\
\text{joint generation:}\quad &P_\theta(V,C\mid I_0,p). \nonumber
\end{align}
Task masks implement all three conditionals with shared weights. To obtain the camera outputs in Fig.~\ref{fig:teaser}, we sample camera latents using the predicted velocities, decode the latents to tipxel maps, and fit camera parameters to those maps.

\subsection{Camera geometry as a three-channel field}
\begin{figure}[t]
\centering
\includegraphics[width=\linewidth]{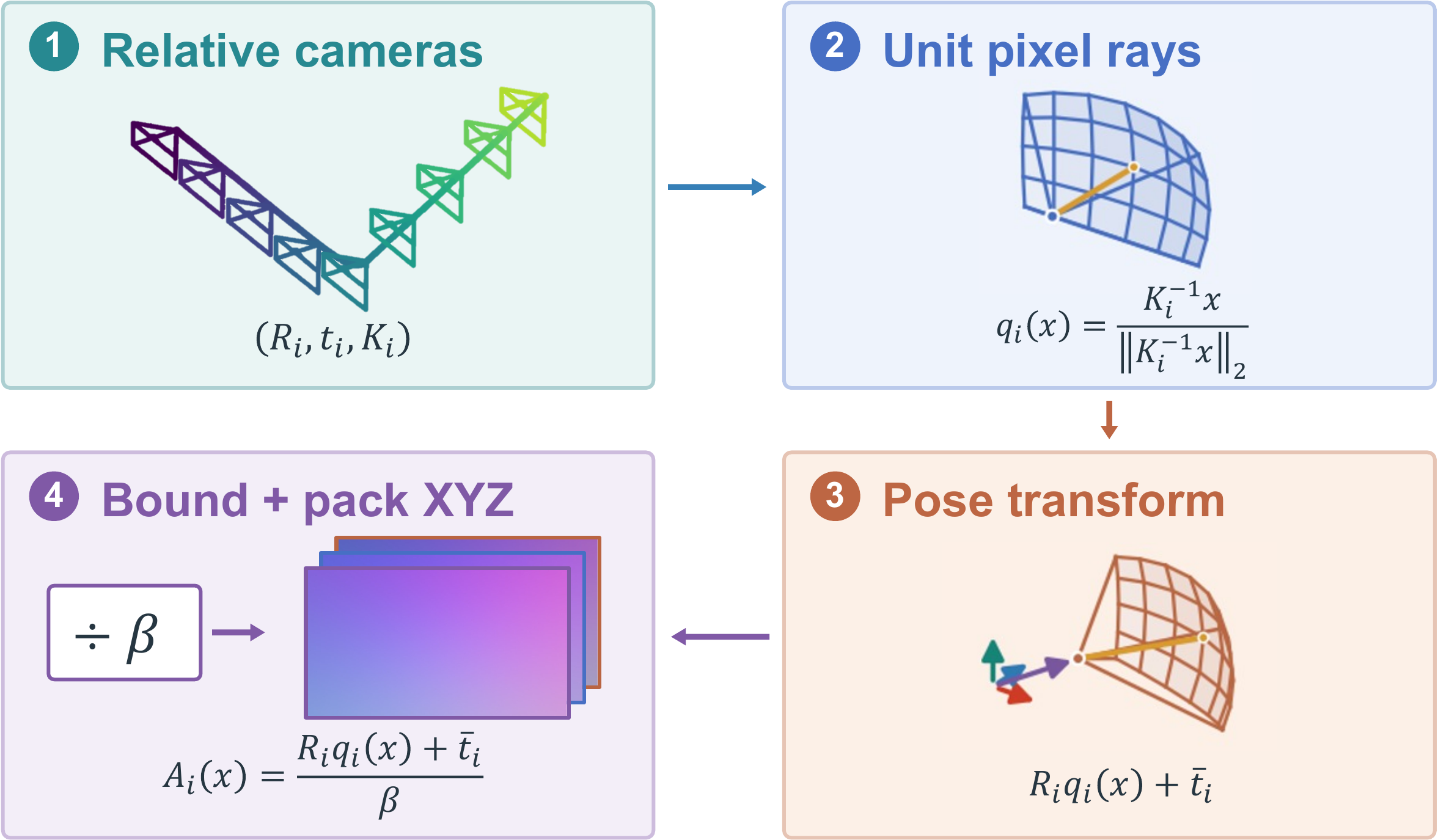}
\caption{\textbf{Encoding cameras as tipxel maps.} Intrinsics back-project homogeneous pixel coordinates $x$ to unit rays. Relative rotation and normalized translation $\bar t_i=t_i/U$ place them in the reference frame; division by $\beta$ bounds the XYZ field. Equation~\ref{eq:tipxel} defines the clip-level constants $U$ and $\beta$. The recovered translations are in normalized clip coordinates.}
\label{fig:camera-encoding}
\end{figure}

A tipxel stores the coordinates of a point one unit along a pixel ray from the normalized camera center. We divide these coordinates by a factor shared across the clip. The three coordinates of each endpoint form the channels of the tipxel map (Fig.~\ref{fig:camera-encoding}). For homogeneous pixel $x=(u,v,1)^\top$, the unit ray in camera coordinates is $q_i(x)=K_i^{-1}x/\|K_i^{-1}x\|_2$. We rotate the ray into the reference frame and normalize the centers and endpoints as follows
\begin{equation}
\begin{aligned}
U&=\max\bigl(1,\max_i\|t_i\|_2\bigr),\qquad \bar t_i=t_i/U,\\
\beta&=1+\max_i\|\bar t_i\|_2,\qquad
A_i(x)=\frac{R_iq_i(x)+\bar t_i}{\beta}.
\end{aligned}
\label{eq:tipxel}
\end{equation}
{\setlength{\emergencystretch}{3em}Since $\|R_iq_i+\bar t_i\|_2\leq\beta$,\linebreak[1] the tipxel map $A_i$ lies in $[-1,1]^{H\times W\times3}$\allowbreak, the video VAE's input range. We visualize the maps with $(A_i+1)/2$ using a fixed mapping across frames. The VAE encodes the floating-point maps directly during training. Since $U$ is not retained, pose estimation and joint generation recover normalized translations. For unknown intrinsics, we assume fixed focal length, a centered principal point, zero skew, and no distortion; known control cameras use the supplied calibration.\par}

\subsection{A unified dual-stream flow model}
\label{sec:model}
\noindent\textbf{Coupled dual-stream architecture.}
\label{sec:architecture}
Following latent diffusion modeling~\cite{ldm}, we use a shared video VAE to encode $V$ and $A$ into appearance latents $z$ and geometry latents $r$. Reference and selected target images are also encoded independently with their frame coordinates preserved. During video generation, the source image supplies clean conditioning tokens and the model samples all output frames, including the first. Frozen UMT5 features~\cite{umt5} are mapped by a shared trainable text projection to context $c$. Both the VAE encoder and decoder remain frozen.

Our diffusion Transformer~\cite{dit} is initialized from Wan2.2-TI2V-5B~\cite{wan22} and has 30 blocks, hidden width 3072, and 24 attention heads. Both streams have separate patch embeddings, time projections, modulation, feed-forward networks, and output heads. Within each block, they share the weights of self-attention and text cross-attention. Each depth has its own attention weights. Training updates all DiT parameters (Fig.~\ref{fig:architecture}).

Let $h_z,h_r$ denote the states after each stream's modulated self-attention and residual update. Our dual-stream cross-attention (DSCA) computes
\begin{equation}
\begin{aligned}
\tilde h_z&=h_z+g_z\odot\mathcal A(N_z(h_z),N_r(h_r)),\\
\tilde h_r&=h_r+g_r\odot\mathcal A(N_r(h_r),N_z(h_z)),
\end{aligned}
\label{eq:dsca}
\end{equation}
where $\mathcal A(a,b)$ uses queries from $a$ and keys/values from $b$. Both directions read the same pre-update pair $(h_z,h_r)$. Their query, key, value, and output projections, as well as query/key normalization, are shared. The non-affine input normalizations and residual gates are separate. Gates $g_z,g_r$ start at zero, initially suppressing the newly introduced exchange while keeping the pretrained within-stream pathways active. The modal projections inherit the pretrained self-attention weights.

Rotary position embeddings~\cite{rope} are applied along aligned three-dimensional positions for self-attention and modal queries/keys. Each block next applies text cross-attention: queries come from the normalized stream state, while keys and values come from the same projected context $c$ in both streams (Fig.~\ref{fig:architecture}b). Shared text-attention weights and stream-specific normalization precede separate modulated feed-forward updates with residual connections. Separate time-conditioned heads after the stack produce $\hat v_z$ and $\hat v_r$.

\noindent\textbf{Masked multi-task flow matching.}
\label{sec:training}
The task determines binary masks $m_z,m_r$, where one denotes an unknown element. Tasks are sampled uniformly. In camera-control training, the reference camera is additionally hidden with probability $0.5$, covering the interface in which only target cameras are supplied.

Following flow matching and rectified flow~\cite{flowmatching,rectifiedflow}, independent Gaussian noise $\epsilon$ defines a linear path for either latent modality $x_0\in\{z_0,r_0\}$:
\begin{equation}
x_\sigma=(1-m)\odot x_0+m\odot[(1-\sigma)x_0+\sigma\epsilon],
\label{eq:path}
\end{equation}
with target velocity $v^*=\epsilon-x_0$ on unknown elements. Observed tokens have noise level zero and receive no loss. We draw a common noise level for both modalities with probability $0.5$; otherwise the two levels are independent. Gaussian noise samples remain independent even when levels match. The configured schedule maps discrete $s\in\{1/1000,\ldots,1\}$ to $\sigma=5s/(1+4s)$.

For each sample, let $a$ and $b$ concatenate the predicted and target velocities over \emph{all unknown elements in both modalities}. The objective is
\begin{equation}
\mathcal L=\frac{\|a-b\|_2^2}{M}+\lambda\bigl(1-\operatorname{cos}(a,b)\bigr),\quad\lambda=0.5,
\label{eq:loss}
\end{equation}
where $M$ counts the unknown elements across both modalities. The squared error gives each element equal weight, and the cosine term measures the alignment of the two velocity vectors with a safeguard for small norms. Camera recovery takes place after decoding, outside the training objective.

\subsection{Conditional sampling and camera recovery}
All tasks use a conditional Euler sampler. Unknown latents start from independent Gaussian noise. For modality $x$ at step $k$,
\begin{equation}
x^{k+1}\leftarrow x^k+(\sigma_x^{k+1}-\sigma_x^k)\hat v_x^k,
\label{eq:euler}
\end{equation}
followed by restoring known latent elements to their clean conditions. We use 50 synchronous steps with guidance scale one, then decode the generated RGB and tipxel latents with the frozen VAE. Algorithm~\ref{alg:sampling} in Appendix~\ref{app:sampling} gives the sampling procedure.

For a decoded field $\hat A_i$, camera recovery fits a rigid transform from unit ray templates to the rescaled field:
\begin{equation}
(\hat R_i,\widehat{\bar t}_i)=\argmin_{R\in\mathrm{SO}(3),t\in\R^3}
\sum_x w_x\|\hat\beta\hat A_i(x)-Rq_i(x)-t\|_2^2.
\label{eq:fit}
\end{equation}
When intrinsics are unknown, we first estimate the fixed focal length and $\beta$ from the reference tipxel map (Appendix~\ref{app:sampling}). We then fit the transform using reweighted rigid SVD~\cite{kabsch}, with a residual weight $w_x$ for each pixel, and align the recovered cameras to the source frame. These camera parameters and normalized trajectories are paired with observed or generated RGB; camera control retains its supplied trajectory.

\begin{figure*}[!t]
\centering
\includegraphics[width=0.98\textwidth]{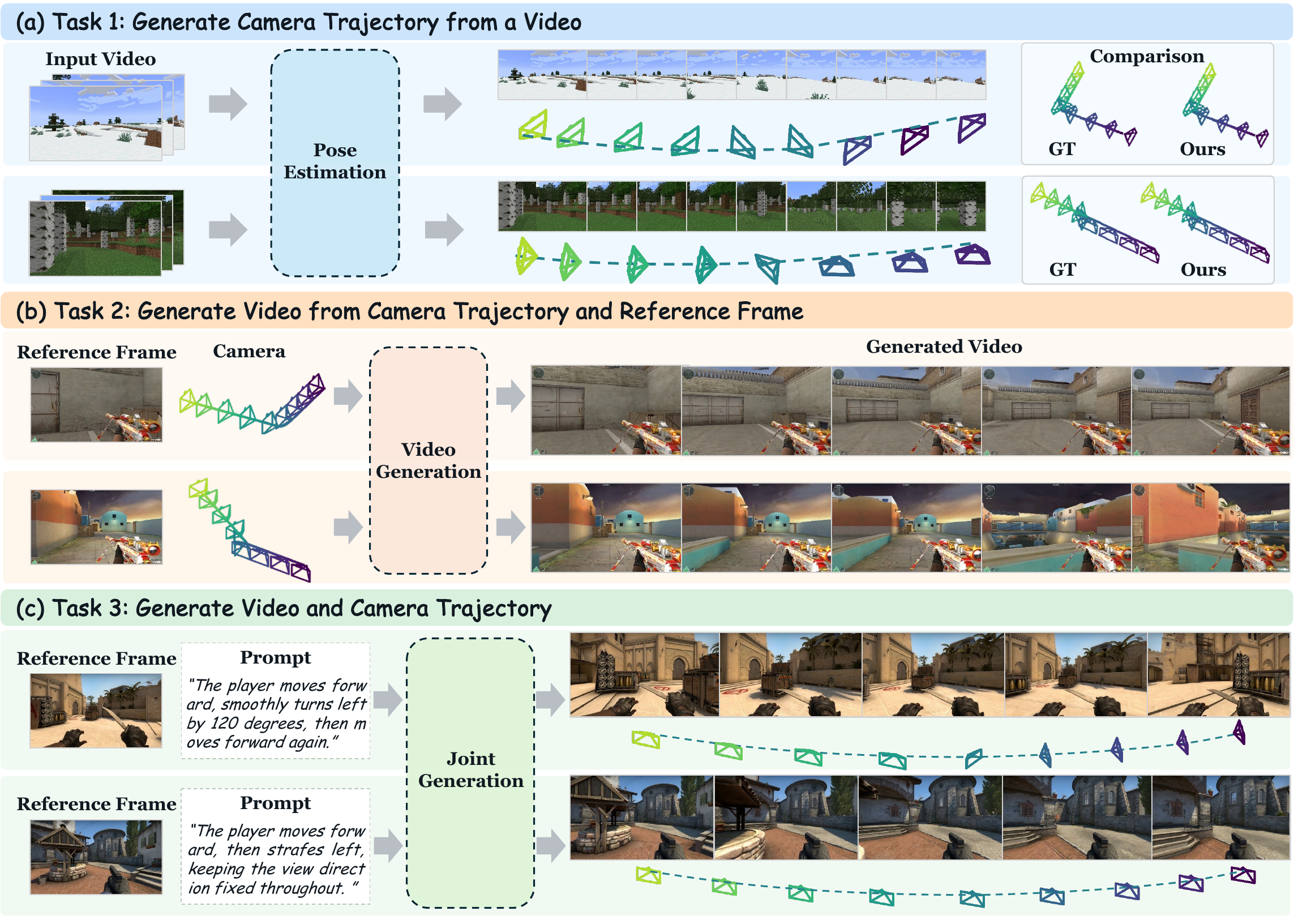}
\caption{\textbf{Qualitative results across tasks.} \textbf{(a)} Pose estimation on Minecraft videos. The right-hand panels compare estimated and ground-truth (GT) cameras. \textbf{(b)} Camera control generates CF videos from reference frames along specified paths. \textbf{(c)} Joint generation produces CS:GO videos and camera trajectories from reference images and text. Frusta show camera poses along each trajectory.}
\label{fig:main-result}
\end{figure*}

\subsection{Data collection and curation}
\label{sec:data}
We retain 100,000 video--camera pairs across four games after screening and manual review (Appendix~\ref{app:collection-pipeline}). The training catalog contains 77,000 of these clips: 10,000 CF, 30,000 CS:GO, 20,000 Minecraft, and 17,000 Arma Reforger clips (Appendix~\ref{app:multi-game}). Each sample pairs 81 RGB frames at $1024\times576$ with rendering-camera extrinsics and intrinsics. Thirteen action families cover translation, turns, jumps, crouches, and ordered combinations (Appendix~\ref{app:collection-pipeline}).

Reliable collection requires synchronizing images and camera states while executing actions despite collisions, terrain, and engine-specific timing. Our backends link CF frames to rendering matrices through shared identifiers, record CS:GO RGB and cameras in the same replay take, synchronize Minecraft render callbacks with simulation and physics updates, and capture Arma camera states at screenshot requests. Geometry-filtered starts and motion checks reduce blocked actions. CF enforces at most 10\,ms capture lateness; CS:GO checks turn-phase alignment within approximately 31.6\,ms (Appendices~\ref{app:crossfire}--\ref{app:arma}).

Automatic validation rejects missing frames, invalid cameras, and inconsistent image--camera associations; manual review checks visual quality and motion-description agreement. Each retained clip includes the frame timestamps and camera matrices. We export these with the action fields and the indices of the selected frames so that each pose can be traced to its source (Appendix~\ref{app:screening}).

\section{Experiments}
\label{sec:experiments}
We compare methods on pose estimation, camera control, and joint generation. Table~\ref{tab:main} gives the quantitative results, and Fig.~\ref{fig:main-result} shows examples from the four-game benchmark. Table~\ref{tab:preference} reports first-choice votes for each task. Tables~\ref{tab:ablations}--\ref{tab:loss-ablation} examine joint sampling and cross-modal attention, noise schedules, and the velocity loss, respectively.

\begin{table*}[t]
\centering
\caption{\textbf{Quantitative results across our three task settings.} CS:GO contributes 40 cases; CF, Minecraft, and Arma Reforger contribute 20 each. Appendix~B defines the metrics and Table~S4 reports pose errors by game. Superscripts rank \method (blue) within each task using unrounded scores. MC: MC-MEt3R; MN: MN-MS; Cam: instruction adherence (0--100).}
\label{tab:main}
\begingroup
\definecolor{gwrow}{HTML}{EAF4FB}
\definecolor{gwaccent}{HTML}{175B83}
\newcommand{\gwrank}[2]{\textbf{#1}\textsuperscript{\textcolor{gwaccent}{\fontsize{6}{6}\selectfont\#\ #2}}}
\fontsize{8.8}{10.5}\selectfont
\setlength{\tabcolsep}{2pt}
\renewcommand{\arraystretch}{1.21}
\begin{minipage}[t]{0.35\textwidth}
\vspace{0pt}
\textbf{(a) Pose estimation}\par\vspace{2pt}
\begin{tabularx}{\linewidth}{@{}>{\hsize=1.48\hsize\linewidth=\hsize\raggedright\arraybackslash}X>{\hsize=0.76\hsize\linewidth=\hsize\raggedleft\arraybackslash}X>{\hsize=0.76\hsize\linewidth=\hsize\raggedleft\arraybackslash}X@{}}
\toprule
Method & tMSE\dn & R-MSE\dn\\\midrule
COLMAP~\cite{colmap} & 3639.32 & 492.85 \\ 
ViPE~\cite{vipe} & 12304.23 & 42.99 \\ 
\shortstack[l]{PoseDiffusion--RE10K\\(GGS-off)~\cite{posediffusion}} & 38279.15 & 2150.23 \\ 
RayDiffusion~\cite{raydiffusion} & 64452.42 & 3140.79 \\ 
DiffusionSfM~\cite{diffusionsfm} & 36160.90 & 1345.97 \\ 
\rowcolor{gwrow}
\textcolor{gwaccent}{\textbf{\method}} & \gwrank{866.25}{1} & \gwrank{345.82}{2} \\ 
\bottomrule
\end{tabularx}
\end{minipage}\hfill
\begin{minipage}[t]{0.63\textwidth}
\vspace{0pt}
\textbf{(c) Joint generation}\par\vspace{2pt}
\begin{tabularx}{\linewidth}{@{}>{\hsize=2.6\hsize\linewidth=\hsize\raggedright\arraybackslash}X>{\hsize=0.68\hsize\linewidth=\hsize\raggedleft\arraybackslash}X>{\hsize=0.68\hsize\linewidth=\hsize\raggedleft\arraybackslash}X>{\hsize=0.68\hsize\linewidth=\hsize\raggedleft\arraybackslash}X>{\hsize=0.68\hsize\linewidth=\hsize\raggedleft\arraybackslash}X>{\hsize=0.68\hsize\linewidth=\hsize\raggedleft\arraybackslash}X@{}}
\toprule
System & IQ\up & AQ\up & MC\dn & MN\up & Cam\up\\\midrule
GenDoP $\rightarrow$ SCoPE~\cite{gendop,scope} & 0.7302 & 0.5388 & 0.1983 & 0.9216 & 60.65 \\ 
Wan2.2-I2V-A14B $\rightarrow$ ViPE~\cite{wan,wan22,vipe} & 0.7362 & 0.5532 & 0.1780 & 0.9278 & 57.63 \\ 
GenDoP $\rightarrow$ GEN3C~\cite{gendop,gen3c} & 0.6870 & 0.4826 & 0.3159 & 0.8631 & 58.96 \\ 
Wan2.2-TI2V-5B $\rightarrow$ ViPE~\cite{wan,wan22,vipe} & 0.6931 & 0.4835 & 0.2139 & 0.9161 & 54.37 \\ 
HunyuanVideo-1.5 I2V $\rightarrow$ ViPE~\cite{hunyuan15,vipe} & 0.7509 & 0.5404 & 0.2613 & 0.9067 & 64.71 \\ 
GenDoP $\rightarrow$ SANA-WM~\cite{gendop,sanawm} & 0.7662 & 0.5307 & 0.3727 & 0.7709 & 52.91 \\ 
\rowcolor{gwrow}
\textcolor{gwaccent}{\textbf{\method}} & \gwrank{0.7509}{3} & \gwrank{0.5432}{2} & \gwrank{0.1427}{1} & \gwrank{0.9450}{1} & \gwrank{85.48}{1} \\ 
\bottomrule
\end{tabularx}
\end{minipage}
\par\vspace{5pt}
\noindent\makebox[\textwidth][l]{\textbf{(b) Camera control}}\par\vspace{2pt}
\begin{tabularx}{\linewidth}{@{}>{\hsize=2.8\hsize\linewidth=\hsize\raggedright\arraybackslash}X>{\hsize=0.7428571428571429\hsize\linewidth=\hsize\raggedleft\arraybackslash}X>{\hsize=0.7428571428571429\hsize\linewidth=\hsize\raggedleft\arraybackslash}X>{\hsize=0.7428571428571429\hsize\linewidth=\hsize\raggedleft\arraybackslash}X>{\hsize=0.7428571428571429\hsize\linewidth=\hsize\raggedleft\arraybackslash}X>{\hsize=0.7428571428571429\hsize\linewidth=\hsize\raggedleft\arraybackslash}X>{\hsize=0.7428571428571429\hsize\linewidth=\hsize\raggedleft\arraybackslash}X>{\hsize=0.7428571428571429\hsize\linewidth=\hsize\raggedleft\arraybackslash}X@{}}
\toprule
Method & ntMSE\dn & R-MSE\dn & IQ\up & AQ\up & MC\dn & MN\up & Cam\up\\\midrule
GEN3C~\cite{gen3c} & 0.0194 & 627.27 & 0.6433 & 0.4688 & 0.1881 & 0.9263 & 76.57 \\ 
SCoPE~\cite{scope} & 0.0127 & 261.65 & 0.7294 & 0.5391 & 0.1367 & 0.9516 & 79.81 \\ 
DualCamCtrl~\cite{dualcam} & 0.0162 & 543.92 & 0.7174 & 0.4959 & 0.1844 & 0.9379 & 75.12 \\ 
WorldStereo (camera-only)~\cite{worldstereo,hyworld} & 0.0151 & 379.01 & 0.7467 & 0.5119 & 0.1254 & 0.9430 & 75.55 \\ 
Uni3C (camera-only)~\cite{uni3c} & 0.0139 & 455.57 & 0.7440 & 0.5343 & 0.2200 & 0.9270 & 81.27 \\ 
SolarWM~\cite{solarwm} & 0.0115 & 47.60 & 0.7251 & 0.5158 & 0.1481 & 0.9389 & 80.98 \\ 
LingBot-World (Cam)~\cite{lingbot} & 0.0048 & 326.85 & 0.7483 & 0.5541 & 0.1470 & 0.9462 & 81.58 \\ 
SANA-WM~\cite{sanawm} & 0.0205 & 330.82 & 0.7491 & 0.5367 & 0.2953 & 0.8754 & 75.68 \\ 
\rowcolor{gwrow}
\textcolor{gwaccent}{\textbf{\method}} & \gwrank{0.0029}{1} & \gwrank{67.09}{2} & \gwrank{0.7508}{1} & \gwrank{0.5426}{2} & \gwrank{0.1257}{2} & \gwrank{0.9518}{1} & \gwrank{86.46}{1} \\ 
\bottomrule
\end{tabularx}
\endgroup
\end{table*}

\subsection{Data, implementation, and protocol}
\label{sec:protocol}
Our 100 evaluation cases include 40 from CS:GO and 20 each from CF, Minecraft, and Arma Reforger. We use the 77,000-clip training catalog in Sec.~\ref{sec:data}. Training uses 24 NVIDIA A800 GPUs, global batch size 24, and AdamW with learning rate $10^{-5}$, weight decay $0.01$, and 400 warm-up updates, for up to 68,000 updates. The Transformer uses BF16 and the frozen VAE FP32. The main benchmark uses 50 Euler steps and guidance scale one. Appendix~\ref{sec:implementation} gives the model settings, Appendix~\ref{sec:training24} describes training, and Appendix~\ref{sec:baselines} lists the baseline configurations.

Pose estimation uses translation MSE (tMSE) after whole-clip Sim(3) alignment~\cite{umeyama} and first-frame-relative rotation MSE (R-MSE, deg$^2$). For camera control, translation MSE is normalized by the requested trajectory's squared maximum displacement (ntMSE), using cameras recovered from generated RGB by DA3-GIANT~\cite{da3}. 

IQ measures image quality with MUSIQ-SPAQ~\cite{musiq}; AQ measures aesthetics with a CLIP-based predictor~\cite{clip}, following VBench~\cite{vbench}. Visual metrics use 11 frames sampled by relative progress within the scoring window; Cam uses the scored video sequence. Scoring covers the first four seconds for CF and five seconds for the other games, or the full output when shorter. Table~\ref{tab:main} averages finite, defined case scores and retains genuine zeros. COLMAP has 99 valid cases for tMSE and 54 for R-MSE; all other cells use 100 cases. Appendix~\ref{app:frame-sampling} gives the frame sampling rule, and Table~\ref{tab:nativeprotocol} lists the native output lengths.

For each input, $r_a=m_a/m_{\rm ref}$ is method $a$'s mean UFM flow magnitude~\cite{ufm} relative to the median flow magnitude of a fixed set of methods on that input. We use separate reference sets for camera control and joint generation (Appendix~\ref{app:motion}). All methods within a task are normalized to the same reference motion. To account for the lower difficulty of matching nearly static frames, we normalize the MEt3R discrepancy $e_a$~\cite{metr} and the AMT-S smoothness deficit $1-MS_a$~\cite{amt} per clip: \begin{equation} \mathrm{MC\!-\!MEt3R}_a=\frac{e_a}{r_a}, \qquad \mathrm{MN\!-\!MS}_a=1-\frac{1-MS_a}{r_a}. \label{eq:motion-calibration} \end{equation}

To score instruction adherence, Qwen3.5-27B uses the generated video and motion cues recovered independently by DA3-GIANT. It checks actions such as turning, jumping, and crouching against the prompt at 21 temporal anchors (Appendix~\ref{app:cam}). The score is $\mathrm{Cam}=100\gamma K^{-1}\sum_{j=1}^K q_j$, where $q_j$ measures adherence to clause $j$ as a binary action check or the fraction of ten satisfied windows. Completing all the requested actions in the wrong order halves the score ($\gamma=0.5$). In all other cases, $\gamma=1$.

\subsection{Pose estimation from observed video}
\method takes the observed video and its action prompt. The baseline estimators receive the observed RGB sequence through their native interfaces (Appendix~\ref{app:baseline-pose}). The pose comparison includes COLMAP~\cite{colmap}, ViPE~\cite{vipe}, PoseDiffusion--RE10K~\cite{posediffusion}, RayDiffusion~\cite{raydiffusion}, and DiffusionSfM~\cite{diffusionsfm}. \method has lower mean translation error than COLMAP in each game (Appendix~\ref{sec:groupresults}, Table~\ref{tab:formalgroup}). In Table~\ref{tab:main}(a), ViPE achieves the lowest rotation MSE ($42.99$\,deg$^2$), followed by \method ($345.82$\,deg$^2$).

\begin{table*}[t]
\centering\small
\caption{\textbf{Human preference across three tasks.} Each column gives first-choice counts and percentages for 60 votes. In (c), agreement measures how well the exported camera trajectory matches the video. Bold marks the highest count in each column, including ties.}
\label{tab:preference}
\definecolor{prefhead}{HTML}{EDF2F7}
\setlength{\tabcolsep}{3pt}
\renewcommand{\arraystretch}{1.14}
\begin{minipage}[t]{0.38\linewidth}
\vspace{0pt}
\begin{tabularx}{\linewidth}{>{\raggedright\arraybackslash}X r}
\toprule
\rowcolor{prefhead}\multicolumn{2}{l}{\textbf{(a) Pose estimation}}\\
Method & First choice\\
\textbf{\method} & \textbf{49 (81.7\%)}\\[5pt]
COLMAP & 9 (15.0\%)\\[5pt]
PoseDiffusion & 0 (0.0\%)\\[5pt]
RayDiffusion & 1 (1.7\%)\\[5pt]
DiffusionSfM & 1 (1.7\%)\\[4pt]
\bottomrule[\lightrulewidth]
\end{tabularx}
\end{minipage}\hfill
\begin{minipage}[t]{0.60\linewidth}
\vspace{0pt}
\begin{tabularx}{\linewidth}{>{\raggedright\arraybackslash}X rr}
\toprule
\rowcolor{prefhead}\multicolumn{3}{l}{\textbf{(b) Camera control}}\\
Method & Trajectory following & Visual quality\\
\textbf{\method} & \textbf{23 (38.3\%)} & \textbf{18 (30.0\%)}\\
GEN3C & 1 (1.7\%) & 0 (0.0\%)\\
SCoPE & 9 (15.0\%) & 1 (1.7\%)\\
DualCamCtrl & 0 (0.0\%) & 0 (0.0\%)\\
WorldStereo & 3 (5.0\%) & 0 (0.0\%)\\
Uni3C & 8 (13.3\%) & 16 (26.7\%)\\
SolarWM & 4 (6.7\%) & 6 (10.0\%)\\
LingBot-World & 9 (15.0\%) & 16 (26.7\%)\\
SANA-WM & 3 (5.0\%) & 3 (5.0\%)\\
\bottomrule[\lightrulewidth]
\end{tabularx}
\end{minipage}\par\vspace{3pt}
\begin{tabularx}{\textwidth}{>{\raggedright\arraybackslash}X rrr}
\toprule[\lightrulewidth]
\rowcolor{prefhead}\textbf{(c) Joint generation} & Agreement & Instruction execution & Visual quality\\
\textbf{\method} & \textbf{40 (66.7\%)} & \textbf{47 (78.3\%)} & \textbf{37 (61.7\%)}\\
GenDoP $\to$ SCoPE & 4 (6.7\%) & 3 (5.0\%) & 4 (6.7\%)\\
Wan2.2-I2V-A14B $\to$ ViPE & 1 (1.7\%) & 1 (1.7\%) & 3 (5.0\%)\\
GenDoP $\to$ GEN3C & 0 (0.0\%) & 0 (0.0\%) & 0 (0.0\%)\\
Wan2.2-TI2V-5B $\to$ ViPE & 6 (10.0\%) & 6 (10.0\%) & 0 (0.0\%)\\
HunyuanVideo-1.5 I2V $\to$ ViPE & 8 (13.3\%) & 3 (5.0\%) & 4 (6.7\%)\\
GenDoP $\to$ SANA-WM & 1 (1.7\%) & 0 (0.0\%) & 12 (20.0\%)\\
\bottomrule
\end{tabularx}
\end{table*}

\subsection{Camera-controlled video generation}
We compare with GEN3C~\cite{gen3c}, SCoPE~\cite{scope}, DualCamCtrl~\cite{dualcam}, WorldStereo~\cite{worldstereo,hyworld}, Uni3C~\cite{uni3c}, SolarWM~\cite{solarwm}, LingBot-World~\cite{lingbot}, and SANA-WM~\cite{sanawm}. All methods receive a reference image and a camera path. We also supply the action prompt to each method that accepts text. WorldStereo and Uni3C use their camera-only configurations. Table~\ref{tab:main}(b) shows that \method achieves the lowest ntMSE ($0.0029$), highest IQ ($0.7508$), highest MN-MS ($0.9518$), and highest Cam ($86.46$). Its ntMSE is $39.6\%$ lower than the next-best value, $0.0048$ from LingBot-World. It ranks second in rotation accuracy, aesthetics, and motion-calibrated consistency. SolarWM has the lowest rotation MSE ($47.60$ versus $67.09$\,deg$^2$ for \method). WorldStereo has the lowest MC-MEt3R ($0.1254$ versus $0.1257$). The two pose metrics capture different aspects of control: ntMSE measures the shape of the aligned path, and R-MSE measures orientation relative to the first frame (Appendix~\ref{app:pose-errors}).

\subsection{Joint video--camera generation}
{\setlength{\emergencystretch}{3em}We compare joint synthesis with two cascade designs. Trajectory-first systems combine GenDoP~\cite{gendop} with SCoPE, GEN3C, or SANA-WM. Video-first systems combine Wan2.2-\allowbreak I2V-A14B, Wan2.2-\allowbreak TI2V-5B~\cite{wan,wan22}, or HunyuanVideo-1.5 I2V~\cite{hunyuan15} with ViPE. All systems receive a reference image and an action prompt. GenDoP supplies cameras to the video generator, while ViPE estimates them from the completed video (Appendix~\ref{app:baseline-joint}). In Table~\ref{tab:main}(c), \method attains Cam $85.48$, exceeding the strongest cascade, HunyuanVideo--ViPE ($64.71$), by $20.77$ points. It also achieves the lowest MC-MEt3R ($0.1427$) and highest MN-MS ($0.9450$), indicating stronger visual consistency and smoother motion under the motion-normalized evaluation. Its IQ is $0.7509$ and AQ is $0.5432$, compared with the respective best values of $0.7662$ and $0.5532$. On 20 fixed Minecraft inputs from the training catalog, average scores varied little across six inference seeds, with mean Cam ranging from $81.76$ to $83.55$ on its 100-point scale (Appendix~\ref{sec:inference-seeds}, Table~\ref{tab:mc_joint_inference_seeds}).\par}

\subsection{Human preference evaluation}
\label{sec:preference}
The study involved 60 participants and recorded 60 first-choice votes for each criterion in Table~\ref{tab:preference}. For pose estimation, \method received the most first-choice votes (49/60), followed by COLMAP (9/60). For camera control, \method led in both trajectory following (23/60) and visual quality (18/60), with Uni3C and LingBot-World tied for second in visual quality (16/60 each). For joint generation, \method led on all three criteria: agreement between the exported camera trajectory and the video (40/60), instruction execution (47/60), and visual quality (37/60).

\begin{table*}[t]
\centering\small
\caption{\textbf{Generation order and cross-modal attention.} Video-first estimates cameras from the generated video. Plain-concat jointly generates video and cameras with concatenated attention in place of dual-stream cross-attention. Bold marks the best value per metric.}
\label{tab:ablations}
\setlength{\tabcolsep}{4pt}
\newcommand{\ablationhead}[2]{\begin{tabular}[c]{@{}c@{}}#1\\#2\end{tabular}}
\begin{tabular*}{\textwidth}{@{\extracolsep{\fill}}*{7}{c}@{}}
\toprule
Variant & \ablationhead{PSNR}{(dB)\up} & SSIM\up & \ablationhead{Position}{RMSE\dn} & \ablationhead{Mean rotation}{error ($^\circ$)\dn} & \ablationhead{Translation direction}{error ($^\circ$)\dn} & \ablationhead{Inference}{time (s)\dn}\\\midrule
Video-first & 18.122 & 0.6510 & 0.2627 & 12.870 & 25.807 & 184.75\\
Plain-concat & 18.171 & 0.6578 & 0.2663 & 12.169 & 24.259 & \textbf{133.55}\\
\textbf{\method (ours)} & \textbf{18.584} & \textbf{0.6661} & \textbf{0.1932} & \textbf{11.560} & \textbf{21.946} & 146.68\\
\bottomrule
\end{tabular*}
\end{table*}

\begin{table}[!t]
\centering\small
\caption{\textbf{Dual-time training and sampling.} Joint generation on 20 Minecraft training clips. Steps count RGB/tipxel updates; all settings use 20 full model calls per clip. PSNR is in dB and time in seconds per clip. Pos: position RMSE; Rot and Dir: mean rotation and translation-direction errors in degrees.}
\label{tab:schedule}
\setlength{\tabcolsep}{1.5pt}
\begin{tabular*}{\columnwidth}{@{\extracolsep{\fill}}lrrrrrrr@{}}
\toprule
$p_{\rm shared}$ & Steps & PSNR\up & SSIM\up & Pos\dn & Rot\dn & Dir\dn & Time\dn\\
\midrule
0.5 & 20/20 & 18.22 & 0.6555 & 0.287 & 11.34 & 26.66 & 146.96 \\
0.5 & 20/10 & 18.23 & 0.6554 & 0.284 & 11.21 & 25.24 & 146.48 \\
\addlinespace[2pt]
1.0 & 20/20 & 18.06 & 0.6555 & 0.317 & 14.65 & 27.67 & 146.20 \\
1.0 & 20/10 & 18.06 & 0.6554 & 0.306 & 13.80 & 26.18 & 146.00 \\
\bottomrule
\end{tabular*}
\end{table}

\begin{table}[!t]
\centering\small
\caption{\textbf{Velocity-loss ablation.} Both variants train for 3,000 updates on the same 20 Minecraft clips. We compare MSE alone with MSE plus cosine distance weighted by 0.5.}
\label{tab:loss-ablation}
\begin{tabular*}{\columnwidth}{@{\extracolsep{\fill}}lrr@{}}
\toprule
Metric & MSE + cosine & MSE only\\
\midrule
\multicolumn{3}{@{}l}{\textit{Pose estimation}}\\
Position RMSE\dn & 0.2868 & \textbf{0.2820}\\
Rotation error ($^\circ$)\dn & \textbf{11.33} & 14.58\\
\addlinespace[2pt]
\multicolumn{3}{@{}l}{\textit{Camera control}}\\
Position RMSE\dn & \textbf{0.0230} & 0.0235\\
Rotation error ($^\circ$)\dn & 1.13 & \textbf{1.10}\\
LPIPS\dn & \textbf{0.3192} & 0.3253\\
\addlinespace[2pt]
\multicolumn{3}{@{}l}{\textit{Joint generation}}\\
Position RMSE\dn & 0.2870 & \textbf{0.2813}\\
Rotation error ($^\circ$)\dn & \textbf{11.34} & 14.60\\
LPIPS\dn & \textbf{0.3199} & 0.3249\\
\bottomrule
\end{tabular*}
\end{table}

\subsection{Joint versus factorized modeling}
\label{sec:measured-ablations}
Table~\ref{tab:ablations} compares \method with a video-first variant that recovers cameras only after RGB generation. PSNR and SSIM measure image fidelity. Camera errors use a common reference frame and translation scale. Position RMSE measures error in the camera centers, mean rotation error is the geodesic angle between orientations, and direction error is the angle between predicted and reference displacements for consecutive frames with nonzero motion. Joint modeling improves PSNR from $18.122$ to $18.584$\,dB and SSIM from $0.6510$ to $0.6661$, while reducing position RMSE by $26.46\%$. Mean rotation error falls from $12.870^\circ$ to $11.560^\circ$, and translation-direction error from $25.807^\circ$ to $21.946^\circ$. Inference takes $146.68$\,s compared with $184.75$\,s, a $20.60\%$ reduction. 

\subsection{Cross-modal fusion}
The plain-concat variant replaces DSCA with concatenated attention while retaining modality identity, time conditioning, and separate prediction heads. In Table~\ref{tab:ablations}, DSCA improves PSNR by $0.413$\,dB and SSIM by $0.0083$, reduces position RMSE by $27.45\%$, and lowers both angular errors. Inference time increases from $133.55$ to $146.68$\,s ($9.83\%$). 

\subsection{Dual-time training and sampling}
\label{sec:schedule}
We train two DSCA variants for 3,000 updates with shared noise-level probabilities of 0.5 and 1.0, then evaluate both on 20 Minecraft training clips. Each schedule makes 20 full-network calls per clip, updating the RGB/tipxel streams 20/20 or 20/10 times. Mixed noise levels give lower position and rotation errors under both schedules (Table~\ref{tab:schedule}). At 20/20 updates, position RMSE falls from $0.317$ to $0.287$ and rotation error from $14.65^\circ$ to $11.34^\circ$. Switching to 20/10 changes PSNR by at most $0.01$\,dB and reduces inference time by only $0.20$--$0.48$\,s per clip.

\subsection{Velocity-loss ablation}
\label{sec:loss-ablation}
Table~\ref{tab:loss-ablation} compares MSE alone with MSE plus cosine distance (weight 0.5) after 3,000 updates. Both runs use the same 20 Minecraft clips and 20/20 synchronous steps. The cosine term reduces mean rotation error from $14.58^\circ$ to $11.33^\circ$ for pose estimation and from $14.60^\circ$ to $11.34^\circ$ for joint generation. LPIPS~\cite{lpips}, averaged over 81 frames and then over clips, falls in both video-generation tasks. MSE alone gives slightly lower position RMSE for estimation and joint generation and lower rotation error for camera control. Both runs use the same DSCA architecture, Wan initialization, and effective batch size of eight. MSE alone uses four GPUs with two gradient accumulation steps, and the cosine variant uses eight GPUs, which can affect sample order and numerical reductions.

\section{Conclusion}
\begingroup
\method uses the same model to annotate gameplay recordings, generate videos along prescribed paths, and jointly generate videos and camera trajectories from action prompts. Once trained on engine captures, it performs all three tasks without querying the game engine. Gated attention connects video and tipxel latents during sampling. Geometric fitting recovers camera intrinsics, orientations, and clip-normalized camera centers. \method achieves lower mean translation error than COLMAP in each game, while ViPE has the lowest pooled rotation MSE. In joint generation, \method leads the evaluated cascades in instruction adherence, motion-calibrated consistency, and motion-normalized smoothness. The ablations show gains in image fidelity and camera accuracy from joint sampling and cross-modal attention. Extending the training set to more games and motion patterns and improving orientation estimates would broaden the coverage and accuracy of camera annotation and support more varied training pairs across diverse gameplay scenarios.
\par\endgroup

\section*{LLM Usage Statement}
OpenAI Codex assisted with reference checking, translation, and language polishing of the main paper and supplementary material. The authors reviewed the resulting edits and take full responsibility for the final content.

\bibliographystyle{ieeenat_fullname}
\bibliography{references}

\onecolumn
\appendix
\setcounter{table}{0}
\renewcommand{\thetable}{S\arabic{table}}
\renewcommand{\theHtable}{appendix.\arabic{table}}
\setcounter{equation}{0}
\renewcommand{\theequation}{S\arabic{equation}}
\renewcommand{\theHequation}{appendix.\arabic{equation}}

\section{Implementation Details}
\label{sec:implementation}
\method initializes its dual-stream Transformer from Wan2.2-TI2V-5B~\cite{wan22}. Both streams and the shared text projection are trainable; the video VAE and UMT5 text encoder~\cite{umt5} are frozen. Table~\ref{tab:recipe} lists the model and inference settings. The optimizer, training budget, and sampling mixture are given in Appendix~\ref{sec:training24}.

\begin{table}[ht]
\centering\small
\caption{Model and inference settings.}
\label{tab:recipe}
\begin{tabular}{@{}p{.49\linewidth}p{.45\linewidth}@{}}
\toprule Setting & Value\\\midrule
Initialization & Wan2.2-TI2V-5B\\
Transformer depth / width / heads & 30 / 3072 / 24\\
Latent channels / patch size & 48 / $(1,2,2)$\\
Frames / image resolution & 81 / $1024\times576$\\
Task sampling probabilities & $1/3$, $1/3$, $1/3$\\
Source camera hidden in control & Probability $0.5$\\
Shared noise-level probability & $0.5$\\
Flow shift / cosine coefficient & 5 / 0.5\\
Transformer / VAE precision & BF16 / FP32\\
Inference / diagnostic steps & 50 / 20\\
Guidance scale / prompt dropout & 1 / 0\\\bottomrule
\end{tabular}
\end{table}

\subsection{Latent layout and conditioning}
The causal VAE encodes an 81-frame clip into 21 temporal latent planes. The reference image and three selected target images are encoded independently, giving 25 planes per modality. Each plane has spatial size $36\times64$; $2\times2$ spatial patching yields $18\times32$ tokens per plane and 14,400 tokens per stream. The reference frame is indexed by zero, and the selected targets by 26, 53, and 80. Rotary temporal positions use the corresponding frame indices divided by four.

RGB and tipxel caches store clean latents. Task masks, Gaussian noise, and noise levels are sampled during training. Both streams condition on the same text. We hold the observed latent elements fixed and supervise the velocity only at unknown elements. Camera-control training hides the source camera with probability $0.5$. Table~\ref{tab:masks} lists the three inference interfaces.

\begin{table}[ht]
\centering\small
\caption{Task inputs and outputs. All tasks use text.}
\label{tab:masks}
\setlength{\tabcolsep}{3pt}
\begin{tabular}{@{}>{\raggedright\arraybackslash}p{.28\linewidth}>{\raggedright\arraybackslash}p{.39\linewidth}>{\raggedright\arraybackslash}p{.27\linewidth}@{}}
\toprule Task & Conditions & Prediction\\\midrule
Pose estimation & RGB video & tipxel field and cameras\\
Camera control & Reference image; target cameras & RGB video\\
Joint generation & Reference image & RGB video and tipxel fields\\\bottomrule
\end{tabular}
\end{table}

\subsection{Sampling and camera recovery}
\label{app:sampling}
Algorithm~\ref{alg:sampling} applies conditional Euler sampling with 50 synchronous steps and guidance scale one. Reference-image tokens remain fixed; the decoded video's first frame is generated by the model. The sampler restores known latent elements after each update and decodes the predicted modalities with the frozen VAE.

For camera-to-world poses $T_i$ and source pose $T_s$, encoding uses $T_s^{-1}T_i$. At the source, Eq.~\eqref{eq:tipxel} gives $A_s(x)=q_s(x)/\beta$. The field has magnitude $1/\beta$ and points along the unit ray at pixel $x$. When intrinsics are unknown, we estimate the clip's fixed focal length and $\beta$ from the decoded reference field, assuming a centered principal point, zero skew, and no distortion. Camera control uses the supplied calibration. In the rigid fit of Eq.~\eqref{eq:fit}, $w_x$ weights the squared residual at pixel $x$. We express the recovered poses relative to the fitted source camera and retain the normalized translation scale.

\begin{algorithm}[!htbp]
\caption{Conditional production of a video--camera pair}
\label{alg:sampling}
\begin{algorithmic}[1]
\Require Conditions, prompt $p$, masks $m_z,m_r$, schedules
\State Encode conditions with frozen VAE
\State $c\gets\operatorname{TextProj}_\theta(\operatorname{UMT5}(p))$
\State Initialize unknown $z,r$ with independent Gaussian noise
\For{$k=0,\ldots,K-1$}
  \State $(\hat v_z,\hat v_r)\gets f_\theta(z,r,\sigma_z^k,\sigma_r^k,c,m_z,m_r)$
  \State Update $x\gets x+(\sigma_x^{k+1}-\sigma_x^k)\hat v_x$ for $x\in\{z,r\}$
  \State Restore all known latent elements to clean conditions
\EndFor
\State Decode target RGB/tipxel with frozen VAE; retain given RGB
\State Fit predicted cameras; retain supplied control cameras
\State \Return video, cameras, intrinsics, scale convention
\end{algorithmic}
\end{algorithm}

\section{Evaluation Metrics}
\label{sec:metrics}
We score frame quality and motion on 11 sampled frames per clip. Camera errors use synchronized poses, and the instruction judge uses the scored video sequence.

\subsection{Native-video sampling}
\label{app:frame-sampling}
For the native frames inside the scoring window, $V=\{I_i\}_{i=0}^{F-1}$, sample
\begin{equation}
k_j=\operatorname{round}\!\left(\frac{j(F-1)}{10}\right),
\quad J_j=I_{k_j},\quad j=0,\ldots,10.
\end{equation}
IQ, AQ, UFM motion, MEt3R, and motion smoothness use these 11 frames. The scoring window is 0--4 seconds for CF and 0--5 seconds for the other groups, including the endpoint; shorter clips use their full native span. Frames retain their native timestamps without interpolation or retiming. DA3-based trajectory comparison uses synchronized native frames, while Cam uses the scored sequence and its motion plots. Table~\ref{tab:nativeprotocol} lists native output formats.

\subsection{Pose and camera-control errors}
\label{app:pose-errors}
\paragraph{Translation MSE for pose estimation.}
Let the predicted and reference camera-to-world poses be
\begin{equation}
\hat T_i=\begin{bmatrix}\hat R_i&\hat{\mathbf c}_i\\0&1\end{bmatrix},
\qquad T_i=\begin{bmatrix}R_i&\mathbf c_i\\0&1\end{bmatrix},
\end{equation}
where $\mathbf c_i$ is the camera center. Let $\mathcal T$ be the valid synchronized frame set and $N=|\mathcal T|$. Fit one positive-scale similarity transform to the entire predicted trajectory:
\begin{equation}
\begin{aligned}
(s^*,Q^*,\mathbf b^*)&=\argmin_{\substack{s>0,\ Q\in SO(3),\\\mathbf b\in\R^3}}
 \sum_{i\in\mathcal T}\|sQ\hat{\mathbf c}_i+\mathbf b-\mathbf c_i\|_2^2,\\
\mathrm{tMSE}&=\frac1N\sum_{i\in\mathcal T}
 \|s^*Q^*\hat{\mathbf c}_i+\mathbf b^*-\mathbf c_i\|_2^2.
\end{aligned}
\end{equation}
tMSE averages squared camera-center distances after a single Sim(3) fit~\cite{umeyama}. The pooled tMSE combines squared game units for CF with squared meters for the other three games. Across the 80 cases measured in meters, weighting the game means in Table~\ref{tab:formalgroup} by case count gives approximately $0.4215$\,m$^2$ for \method and $0.4995$\,m$^2$ for COLMAP.

\paragraph{Normalized translation MSE for camera control.}
DA3-GIANT~\cite{da3} recovers the generated video's trajectory. After matching it to the generator's control path, we compute whole-clip Sim(3)-aligned tMSE and normalize by the path's maximum displacement:
\begin{equation}
D_{\max}=\max_i\|\mathbf c_i-\mathbf c_0\|_2,\qquad
\mathrm{ntMSE}=\frac{\mathrm{tMSE}}{D_{\max}^2}.
\label{eq:ntmse}
\end{equation}
For $D_{\max}>0$, ntMSE measures how closely the aligned trajectory follows the shape of the requested path. Dividing by $D_{\max}^2$ makes the score dimensionless.

Control cameras are expressed in the source frame, with translation divided by $\max(1,\max_i\|t_i\|_2)$ over the complete 81-frame path. Rotations, intrinsics, and timestamps are retained. GEN3C and DualCamCtrl use their native-length control paths; WorldStereo uses source indices $0,4,\ldots,80$, and SANA-WM uses $1,\ldots,80$. For SolarWM, the reference is the full 81-frame path. We match the generated frames to the requested cameras by timestamp.

\paragraph{First-frame-relative rotation MSE.}
Convert each sequence to its own initial orientation:
\begin{equation}
\hat R_i^{\mathrm{rel}}=\hat R_0^\top\hat R_i,\qquad
R_i^{\mathrm{rel}}=R_0^\top R_i.
\end{equation}
Compute the geodesic angle in degrees and its mean square:
\begin{equation}
\begin{aligned}
u_i&=\frac{\operatorname{tr}[(\hat R_i^{\mathrm{rel}})^\top
             R_i^{\mathrm{rel}}]-1}{2},\\
\theta_i&=\frac{180}{\pi}\arccos[\operatorname{clip}(u_i,-1,1)],\\
\mathrm{R\!-\!MSE}&=\frac1N\sum_{i\in\mathcal T}\theta_i^2.
\end{aligned}
\end{equation}
We report rotation MSE in deg$^2$ for both pose estimation and camera control. Each rotation sequence uses its own initial orientation, independently of the similarity fit to camera centers.

\subsection{Frame quality}
\paragraph{Imaging quality (IQ).}
VBench v0.1.5's MUSIQ-SPAQ model~\cite{musiq,vbench} scores the 11 sampled frames:
\begin{equation}
\mathrm{IQ}(V)=\frac1{11}\sum_{j=0}^{10}
\frac{q_{\mathrm{MUSIQ}}(P_{\mathrm{IQ}}(J_j))}{100}.
\end{equation}
$P_{\mathrm{IQ}}$ preserves aspect ratio, resizes the long side to 512 only when it exceeds 512, and divides RGB values by 255. Higher IQ indicates better single-frame technical quality.

\paragraph{Aesthetic quality (AQ).}
Use CLIP ViT-L/14~\cite{clip} and the LAION linear aesthetic head:
\begin{equation}
\begin{aligned}
\mathbf z_j&=\frac{f_{\mathrm{CLIP}}(P_{\mathrm{AQ}}(J_j))}
 {\|f_{\mathrm{CLIP}}(P_{\mathrm{AQ}}(J_j))\|_2},\\
\mathrm{AQ}(V)&=\frac1{11}\sum_{j=0}^{10}
 \frac{\mathbf w^\top\mathbf z_j+b}{10}.
\end{aligned}
\end{equation}
$P_{\mathrm{AQ}}$ resizes the short side to 224, center-crops to $224\times224$, and uses official CLIP pixel normalization. The head is \texttt{sa\_0\_4\_vit\_l\_14\_linear}, acting on 768-dimensional normalized features. Higher AQ indicates greater predicted aesthetic quality.

\subsection{Motion-calibrated consistency and smoothness}
\label{app:motion}
\paragraph{UFM apparent motion.}
Bilinearly resize the sampled frames to $W\times H=448\times256$. UFM-Base~\cite{ufm} estimates forward and backward optical flow for ten adjacent pairs, yielding 20 directional fields:
\begin{equation}
m(V)=\frac1{20HW}\sum_{j=0}^{9}
 \sum_{d\in\{\rightarrow,\leftarrow\}}\sum_p
 \|\mathbf f_j^d(p)\|_2.
\end{equation}
$m(V)$ is the all-pixel mean flow magnitude on a fixed grid, measured in pixels.

\paragraph{Fixed reference sets.}
For each task and source condition, form a reference median over a fixed method set $\mathcal C$:
\begin{equation}
m_{\mathrm{ref}}=\operatorname{median}_{a\in\mathcal C}m_a,
\qquad r_a=\frac{m_a}{m_{\mathrm{ref}}}.
\end{equation}
For camera control, $\mathcal C$ contains \method, SCoPE, DualCamCtrl, WorldStereo, Uni3C, and GEN3C. For joint generation, it contains \method, GenDoP--SCoPE, GenDoP--GEN3C, Wan2.2-TI2V-5B--ViPE, and Wan2.2-I2V-A14B--ViPE. We keep each reference set fixed when scoring additional methods. Motion-normalized scores require positive reference motion and $r_a>0$.

\paragraph{MC-MEt3R.}
Bilinearly resize each of the ten adjacent sampled frame pairs to $256\times256$. The MEt3R implementation~\cite{metr} uses MASt3R geometry~\cite{mast3r}, DINO16 features, and FeatUp upsampling. Predicted point maps project the two feature maps into a common coordinate system. For projected features $\phi_a(p),\phi_b(p)$ and overlap mask $M(p)$, define
\begin{equation}
\begin{aligned}
d(p)&=1-\frac{\phi_a(p)^\top\phi_b(p)}
 {\|\phi_a(p)\|_2\|\phi_b(p)\|_2+\epsilon},\\
e_{ab}&=\frac{\sum_pM(p)d(p)}{\sum_pM(p)+\epsilon},
\qquad \epsilon=10^{-5}.
\end{aligned}
\end{equation}
Let $\mathcal P$ contain adjacent pairs with finite scores and positive overlap. The raw and calibrated scores are
\begin{equation}
e(V)=\frac1{|\mathcal P|}\sum_{(a,b)\in\mathcal P}e_{ab},
\qquad \mathrm{MC\!-\!MEt3R}_a=\frac{e_a}{r_a}.
\end{equation}
Lower MC-MEt3R indicates better multi-view consistency relative to apparent motion.

\paragraph{MN-MS.}
VBench's AMT-S~\cite{amt} predicts five intermediate frames from even-indexed sampled frames:
\begin{equation}
\hat J_{2j+1}=f_{\mathrm{AMT-S}}(J_{2j},J_{2j+2};0.5),
\quad j=0,\ldots,4.
\end{equation}
For RGB values on the 0--255 scale, the reconstruction error and raw smoothness are
\begin{equation}
\begin{aligned}
D(V)&=\frac1{5\cdot3HW}
 \sum_{j=0}^{4}\sum_{p,c}|J_{2j+1}(p,c)-\hat J_{2j+1}(p,c)|,\\
MS(V)&=1-\frac{D(V)}{255}.
\end{aligned}
\end{equation}
Use the same relative-motion ratio to define
\begin{equation}
\mathrm{MN\!-\!MS}_a=1-\frac{1-MS_a}{r_a}.
\end{equation}
Higher MN-MS indicates smoother motion relative to apparent motion. Both calibrated scores are computed per clip before averaging and are retained without clipping.

\subsection{Camera instruction adherence (Cam)}
\label{app:cam}
Qwen3.5-27B judges motion clauses using the scored RGB sequence, 21 equally spaced temporal anchors, and motion cues recovered independently by DA3-GIANT. Auxiliary plots use speed range 0--50 and yaw range $-360^\circ$ to $360^\circ$, with ten temporal windows. The same configuration applies to all 1,600 judgments. Clauses cover translation, turning, jumping, and crouching.

Split the instruction into $K$ clauses. An action or count clause receives a Boolean value, while a throughout-the-video clause receives the fraction of satisfied windows:
\begin{equation}
q_j=\begin{cases}
\mathbf1[\text{clause satisfied}],&\text{action/count},\\
\frac1{10}\sum_{w=1}^{10}\mathbf1[\text{window }w\text{ satisfied}],
 &\text{throughout}.
\end{cases}
\end{equation}
The deterministic aggregator computes
\begin{equation}
\mathrm{Cam}=100\left(\frac1K\sum_{j=1}^Kq_j\right)\gamma.
\end{equation}
Here $\gamma=0.5$ when every ordered clause is satisfied but their order is incorrect; otherwise $\gamma=1$. Each clause receives equal weight.

The judge supplies the clause assessments and the order verdict used in the Cam score. Decoding uses seed 42, temperature 0.7, top-$p$ 0.8, top-$k$ 20, presence penalty 1.5, a 3,072-token limit, and disabled thinking. A JSON formatting failure permits one same-input retry with seed 43. A budget-format error permits a forced-JSON retry at the original seed and token limit. The first valid response is retained.

Cam ranges from 0 to 100 and measures instruction adherence visible in RGB.

\subsection{Aggregation and evaluation coverage}
For each task, method and metric, let $\mathcal V_a$ contain cases with finite, defined values. The main table reports
\begin{equation}
\bar s_a=\frac{1}{|\mathcal V_a|}\sum_{n\in\mathcal V_a}s_{a,n},
\qquad n_a=|\mathcal V_a|.
\end{equation}
Genuine zero scores are included. COLMAP has 99 valid cases for tMSE and 54 for R-MSE. All other main-table cells use all 100 cases. Each game contributes in proportion to its valid case count for that metric (Table~\ref{tab:formalgroup}).

IQ, AQ, and motion smoothness use VBench v0.1.5~\cite{vbench}; MEt3R uses implementation revision \texttt{ee0e175}~\cite{metr}.

\section{Baseline Configurations}
\label{sec:baselines}
\subsection{Pose estimation}
\label{app:baseline-pose}
COLMAP~\cite{colmap}, ViPE~\cite{vipe}, PoseDiffusion~\cite{posediffusion}, RayDiffusion~\cite{raydiffusion}, and DiffusionSfM~\cite{diffusionsfm} receive the observed RGB sequence through their native interfaces. PoseDiffusion uses its RE10K checkpoint with geometry-guided sampling disabled. RayDiffusion uses diffusion inference, and DiffusionSfM uses its main model. \method additionally receives the supplied prompt. Translation and rotation are evaluated with the conventions in Appendix~\ref{sec:metrics}.

\subsection{Camera-controlled generation}
All systems receive a source image and target cameras. Methods that accept text also receive the action prompt. Uni3C~\cite{uni3c} and WorldStereo~\cite{worldstereo,hyworld} use camera-only control. DualCamCtrl~\cite{dualcam} receives depth predicted from the source image. SCoPE~\cite{scope} uses its Wan2.2-I2V-A14B release. SolarWM~\cite{solarwm} uses its released normalized intrinsics and causal camera indices $0,1,5,\ldots,77$.

SANA-WM~\cite{sanawm} includes its video refiner. Refinement removes the clean first-frame anchor from an 81-frame input; the resulting 80 frames correspond to source camera indices 1--80, with a $0.0625$-second starting offset. Table~\ref{tab:nativeprotocol} summarizes the outputs of all evaluated generation systems.

\subsection{Joint generation}
\label{app:baseline-joint}
Trajectory-first systems combine GenDoP~\cite{gendop} with SCoPE, GEN3C~\cite{gen3c}, or the full SANA-WM pipeline. GenDoP receives the reference image, text, and source-estimated depth, then supplies predicted cameras and intrinsics to the video generator. Video-first systems combine Wan2.2-I2V-A14B, Wan2.2-TI2V-5B~\cite{wan,wan22}, or HunyuanVideo-1.5~\cite{hunyuan15} with ViPE. HunyuanVideo uses its 480P-I2V pipeline. Every cascade starts from a reference image and text, without a ground-truth target trajectory. Inference proceeds once through the two stages, without pose feedback into video generation.

\begin{table}[tbp]
\centering\small
\caption{Native output formats. Span equals $(\mathrm{frames}-1)/\mathrm{fps}$. \textsuperscript{*}Arma Reforger outputs span 5.33 seconds at 15 fps.}
\label{tab:nativeprotocol}
\setlength{\tabcolsep}{5pt}
\begin{tabular*}{\textwidth}{@{\extracolsep{\fill}}llrrr@{}}
\toprule Task & System & Frames & fps & Span (s)\\\midrule
Camera control & \method\textsuperscript{*} & 81 & 16 & 5.00\\
 & GEN3C & 121 & 24 & 5.00\\
 & SCoPE & 81 & 16 & 5.00\\
 & DualCamCtrl & 61 & 10 & 6.00\\
 & WorldStereo (camera-only) & 21 & 16 & 1.25\\
 & Uni3C (camera-only) & 81 & 16 & 5.00\\
 & SolarWM & 81 & 16 & 5.00\\
 & LingBot-World (Cam) & 81 & 16 & 5.00\\
 & SANA-WM & 80 & 16 & 4.94\\\midrule
Joint generation & \method\textsuperscript{*} & 81 & 16 & 5.00\\
 & GenDoP $\rightarrow$ SCoPE & 81 & 16 & 5.00\\
 & Wan2.2-I2V-A14B $\rightarrow$ ViPE & 81 & 16 & 5.00\\
 & GenDoP $\rightarrow$ GEN3C & 121 & 24 & 5.00\\
 & Wan2.2-TI2V-5B $\rightarrow$ ViPE & 81 & 24 & 3.33\\
 & HunyuanVideo-1.5 I2V $\rightarrow$ ViPE & 121 & 24 & 5.00\\
 & GenDoP $\rightarrow$ SANA-WM & 80 & 16 & 4.94\\\bottomrule
\end{tabular*}
\end{table}

\section{Results by Game}
\label{sec:groupresults}
The benchmark contains 40 CS:GO cases and 20 each from CF, Minecraft, and Arma Reforger. Evaluating six pose-estimation, nine camera-control, and seven joint-generation configurations yields 2,200 records. Table~\ref{tab:formalgroup} gives pose errors and valid case counts for each game under the protocol in Appendix~\ref{sec:metrics}.

\begin{table}[tbp]
\centering\small
\caption{\textbf{Pose estimation by game.} Translation MSE uses squared game units for CF and m$^2$ elsewhere. Rotation MSE uses deg$^2$. CS:GO values combine two rounded group means weighted by valid counts for each metric. All methods shown cover 40 CS:GO cases and 20 per other game except COLMAP, whose counts appear below. Main-paper Table~1(a) reports the pooled ViPE scores on all 100 cases.}
\label{tab:formalgroup}
\setlength{\tabcolsep}{4pt}
\begin{tabular*}{\textwidth}{@{\extracolsep{\fill}}lrrrr@{}}
\toprule\multicolumn{5}{l}{\textbf{Translation MSE}}\\
Method & CF & Arma Reforger & CS:GO & Minecraft\\\midrule
COLMAP & 1.896e+04 & 0.02577 & 0.3546 & 1.263\\
PoseDiffusion & 1.914e+05 & 1.202 & 6.0000 & 7.848\\
RayDiffusion & 3.222e+05 & 1.94 & 7.9540 & 7.691\\
DiffusionSfM (main) & 1.808e+05 & 0.5101 & 3.9345 & 5.126\\
\textbf{\method (ours)} & 4330 & 0.003183 & 0.3034 & 1.076\\
\midrule COLMAP valid $n$ & 19/20 & 20/20 & 40/40 & 20/20\\
\bottomrule\end{tabular*}\par\smallskip
\begin{tabular*}{\textwidth}{@{\extracolsep{\fill}}lrrrr@{}}
\toprule\multicolumn{5}{l}{\textbf{Rotation MSE}}\\
Method & CF & Arma Reforger & CS:GO & Minecraft\\\midrule
COLMAP & 1139 & 55.95 & 134.6 & 1805\\
PoseDiffusion & 1303 & 150.1 & 2921.0 & 3456\\
RayDiffusion & 941.6 & 1129 & 5001.5 & 3630\\
DiffusionSfM (main) & 1518 & 53.67 & 931.4 & 3296\\
\textbf{\method (ours)} & 377.5 & 0.4655 & 224.9 & 901.3\\
\midrule COLMAP valid $n$ & 5/20 & 9/20 & 31/40 & 9/20\\
\bottomrule\end{tabular*}\par\smallskip
\end{table}

\section{Data Collection and Curation}
\label{sec:collection}
\subsection{Collection pipeline}
\label{app:collection-pipeline}
Automatic screening and manual review retained 100,000 video--camera pairs from a larger pool of candidate clips. The four backends cover CrossFire (CF), CS:GO, Minecraft, and Arma Reforger. Each pipeline selects an action and initial state, executes the action, records RGB and rendering-camera states, and validates the exported sequence. Table~\ref{tab:captureprotocol} summarizes synchronization and coordinate conventions.

The action library comprises 13 families covering forward/backward motion, strafing, turns, jumps, crouches, and ordered combinations such as move--turn--move and jump--turn--move. CF, CS:GO, and Minecraft construct descriptions from structured action fields. The camera labels come from the rendering state for each recorded frame.

\begin{table}[tbp]
\centering\small
\caption{Capture protocols for 81 paired RGB frames ($1024\times576$) and cameras. Local axes point right, down, and forward (OpenCV).}
\label{tab:captureprotocol}
\setlength{\tabcolsep}{3pt}
\begin{tabular}{@{}>{\raggedright\arraybackslash}p{.10\textwidth}>{\raggedright\arraybackslash}p{.23\textwidth}>{\raggedright\arraybackslash}p{.31\textwidth}>{\raggedright\arraybackslash}p{.12\textwidth}>{\raggedright\arraybackslash}p{.19\textwidth}@{}}
\toprule Game & Action control and starts & RGB--camera synchronization & Sampling & World coordinates\\\midrule
CF & Native inputs; geometry-filtered starts and headings & D3D9 matrices linked to displayed frames by update and presentation identifiers & 16 Hz; 5 s span & Z up; native game units\\
CS:GO & Server-controlled bot; map anchors & SourceTV replay with RGB and HLAE camera from the same take & 16 fps; 5 s span & Source axes; $0.0254$ m/HU\\
Minecraft & Collision-aware displacement; terrain-filtered routes & Matched render callbacks for framebuffer, eye pose, and projection & 16 Hz; 5 s span & Y up; 1 block = 1 m\\
\shortstack[l]{Arma\\Reforger} & Character controller; 40 spawn points & Camera state and viewport rays recorded at screenshot requests & 15 Hz; $80/15$ s span & Y up; meters\\\bottomrule
\end{tabular}
\end{table}

\subsection{CrossFire}
\label{app:crossfire}
A structured action plan is fixed before start selection. Turn magnitudes are $60^\circ$, $120^\circ$, and $180^\circ$, sampled with probabilities 0.4, 0.4, and 0.2. A motion predictor checks candidate starts and headings at $15^\circ$ increments against collision geometry, body support, and clearance. Native inputs execute five seconds of action followed by 0.5 seconds of released-input observation.

The recorder associates input updates with D3D9 view/projection matrices, viewport parameters, and displayed RGB frames using shared identifiers and a high-resolution clock. At $t_j=t_0+j/16$, it selects the first displayed frame at or after the target time. Acceptance requires 81 distinct frames, at most 10 ms of lateness, at most 1\% mean cadence error, and complete frame-buffer records. Screening checks motion, body clearance, jump/crouch events, cameras, and media. Camera comparisons use at least 16 observations, with position residual at most 0.02 native units and rotation residual at most $0.1^\circ$. Extrinsics derive from view matrices and intrinsics from projection and viewport parameters. World coordinates change from $(X,Y,Z)$ to $(X,Z,Y)$; translation retains native units.

\subsection{CS:GO}
A SourceMod plugin controls a bot in CS:GO Legacy at 128 simulation ticks per second. Each action lasts five seconds, followed by at least 0.5 seconds of final-heading hold. Starts use map anchors with geometry and settling checks. Directions, turn magnitudes, and weapons vary within each action family; turns follow the CF distribution. Runtime checks track commands, eye direction, body motion, support, and progress. Requested jumps require takeoff within 0.03125 seconds.

SourceTV records demo segments and action events. HLAE replays the selected bot in first person, recording RGB, a compressed radial-depth visualization, and the final rendering camera in the same take. A common consecutive 81-frame window is selected across streams. For measurable-turn sequences, alignment searches $\pm2$ frames around the nominal crop and requires phase error at most 0.031640625 seconds and an all-frame yaw-envelope check. Rotations convert from Source roll/pitch/yaw to OpenCV axes; translations use $0.0254$ m/HU. Intrinsics derive from the recorded horizontal field of view.

\subsection{Minecraft}
Minecraft Java 1.21.1 uses a Fabric bridge in desert, forest, and snowy-plains worlds. A $96\times96$-block region is scanned at 0.25-block spacing. Admission requires walkable fraction at least 0.30 and water fraction at most 0.01. Routes retain a 12-block boundary margin, 1.8 blocks of headroom, no water crossings, and ground-height changes at most 0.25 blocks, checked at 0.125-block spacing.

The collector freezes ordinary ticks and advances observations at 16 Hz in simulation time. Horizontal motion uses collision-aware displacement, while jump and gravity follow 20 Hz physics:
\begin{equation}
n_j=\left\lfloor20j/16\right\rfloor,\qquad j=0,\ldots,80.
\end{equation}
Thus, 80 observation intervals contain 100 physics steps. Turns range from $20^\circ$ to $180^\circ$ in $10^\circ$ increments.

Before-render callbacks advance actions; after-render callbacks save RGB, eye pose, and projection with matching frame identifiers. The next step waits for image writing. Acceptance requires 81 synchronized pairs, finite camera matrices, yaw/pitch agreement within $0.001^\circ$, and the requested jump/crouch effects. Quaternions convert from OpenGL to OpenCV axes, retaining Y-up world coordinates. Intrinsics use the actual rendering projection. Simulation, physics, and render timestamps are stored separately.

\subsection{Arma Reforger}
\label{app:arma}
A first-person infantry controller executes 13 action templates at 40 registered spawn points. Spawn slots cycle through the pool, and initial headings are uniform over $[0^\circ,360^\circ)$. The capture configuration uses $45^\circ$ right turns and right strafing. Failed slots retry the same action and spawn with a new heading, for at most ten attempts.

After an eight-second warm-up, 81 screenshots are requested at $t_j=j/15$. Each request records camera state, vertical field of view, projection origin, viewport rays, engine frame identifier, and time. Each screenshot is written before the next request. Identifiers and request times increase monotonically; relative grid error is bounded by $1/15$ second. Camera axes are orthogonalized from viewport rays with a consistent handedness conversion. Translation uses meters and Y-up world coordinates.

Motion checks use 0.5-second and one-second windows to reject stalls, direction errors, and ineffective oscillation, alongside camera, timestamp, and media checks. Captions encode controller intent; fixed templates do not resolve all per-clip direction, angle, and crouch variations. Individual jump, landing, and crouch events are not independently verified by this backend.

\subsection{Screening and export}
\label{app:screening}
Each sample stores RGB video, 81 camera-to-world matrices, 81 intrinsic matrices, timestamps, action fields, and source metadata. CF, CS:GO, and Minecraft use 16 fps with a five-second first-to-last-frame span; Arma uses 15 fps with a $5.333$-second span. Samples with missing observations are rejected without frame duplication or interpolation.

Export validation checks image dimensions, finite cameras, sequence length, and frame--camera correspondence. Manual review assesses visual quality and motion-description agreement. Source identities, selected indices, and collector versions accompany each sample. Model inputs express poses relative to the reference camera, retaining the coordinate and unit conventions in Table~\ref{tab:captureprotocol}.

\section{Training Configuration}
\label{sec:training24}
\subsection{Optimization}
The distributed configuration uses 24 NVIDIA A800 GPUs across three nodes, with eight GPU ranks per node. FSDP applies full parameter sharding and transformer-block wrapping with BF16 computation and FP32 master weights. Each GPU processes one sample per update without gradient accumulation, giving a global batch of 24. AdamW uses learning rate $10^{-5}$, weight decay 0.01, 400 warm-up updates, and gradient clipping at norm one. The maximum update count is 68,000 (Table~\ref{tab:training24budget}).

\begin{table}[t]
\centering\small
\caption{Mixed-training configuration.}
\label{tab:training24budget}
\begin{tabular}{@{}p{.54\linewidth}p{.40\linewidth}@{}}
\toprule Setting & Value\\\midrule
Nodes / GPUs per node & 3 / 8\\
Global / per-GPU batch & 24 / 1\\
Gradient accumulation & 1\\
Optimizer updates & 68,000\\
Learning rate / weight decay & $10^{-5}$ / 0.01\\
Warm-up updates & 400\\
Gradient clipping & Norm 1\\
Sample presentations & 1,632,000\\
Virtual epochs & 16\\
Checkpoint / diagnostic interval & 5,000 updates\\
Diagnostic sampling steps & 20\\\bottomrule
\end{tabular}
\end{table}

\subsection{Multi-game sampling}
\label{app:multi-game}
The active training catalog contains 77,000 records: 10,000 from CF, 17,000 from Arma Reforger, 30,000 from the merged CS:GO collection, and 20,000 from Minecraft. A virtual epoch has 102,000 sample slots in the ratio $15{,}000:15{,}000:36{,}000:36{,}000$. Samples from different games are mixed within global batches. Table~\ref{tab:training24mixture} gives source counts and exposure.

Each rank processes 4,250 slots per virtual epoch. Consequently, 68,000 updates correspond to 16 virtual epochs and $68{,}000\times24=1{,}632{,}000$ sample presentations. The per-game quotas count presentations sampled from the 77,000-clip training subset of the 100,000-pair collection.

\begin{table}[htb]
\centering\small
\caption{Source mixture and sample exposure. Presentations include repeated sampling over 16 virtual epochs.}
\label{tab:training24mixture}
\setlength{\tabcolsep}{3pt}
\fontsize{8.5}{10}\selectfont
\begin{tabular*}{\linewidth}{@{\extracolsep{\fill}}lrrr@{}}
\toprule Source & Records & Slots/epoch & Presentations\\\midrule
CF & 10,000 & 15,000 & 240,000\\
Arma Reforger & 17,000 & 15,000 & 240,000\\
CS:GO (merged) & 30,000 & 36,000 & 576,000\\
Minecraft & 20,000 & 36,000 & 576,000\\\midrule
Total & 77,000 & 102,000 & 1,632,000\\\bottomrule
\end{tabular*}
\end{table}

\subsection{Execution and diagnostics}
RGB and tipxel VAE caches are reused across epochs. The runtime uses PyTorch 2.13.0+cu130, CUDA 13.0, and Accelerate 1.14.0. All ranks participate in FSDP; rank zero writes checkpoints. Checkpoint and diagnostic intervals are 5,000 updates, with an additional final checkpoint at update 68,000. This gives 14 evaluation points across the training sources. Diagnostics use 20 Euler steps and cover the three tasks, loop consistency, XYZ fields, and yaw/pitch/roll. These diagnostics use the training sources. Appendix~\ref{sec:metrics} gives the separate protocol for the 100-case benchmark.

\section{Inference-Seed Sensitivity}
\label{sec:inference-seeds}
We varied the initial sampling noise for joint generation while fixing the Minecraft checkpoint at training step 15,000. For each seed, we used 50 synchronous steps and guidance scale one to generate a video and camera trajectory for each of the same 20 inputs from the training catalog.

\begin{table}[tbp]
\centering
\caption{Inference-seed sensitivity for joint generation on the 20 fixed Minecraft first-frame/prompt pairs from the training catalog (17 distinct texts). We use the same 15k checkpoint and 50 sampling steps throughout and reuse the original outputs for seed 42. Each column under inference seed averages the same \(n\) inputs with valid scores at all six seeds. SD is the sample standard deviation of the six means. The final two columns give the range of these means and the median of the per-input ranges. Cam uses its 0--100 scale.}
\label{tab:mc_joint_inference_seeds}
\small
\setlength{\tabcolsep}{3.4pt}
\begin{tabular}{lrrrrrrrrrr}
\toprule
 & & \multicolumn{6}{c}{Inference seed} & \multicolumn{2}{c}{Seed means} & Within input \\
\cmidrule(lr){3-8}\cmidrule(lr){9-10}
Metric & \(n\) & 0 & 1 & 10 & 42 & 100 & 1000 & SD & Range & Median range \\
\midrule
MC-MEt3R $\downarrow$ & 20 & 0.1168 & 0.1019 & 0.1123 & 0.1072 & 0.1234 & 0.1149 & 0.0075 & 0.0214 & 0.0570 \\
MN-MS $\uparrow$ & 20 & 0.9567 & 0.9629 & 0.9610 & 0.9611 & 0.9572 & 0.9589 & 0.0024 & 0.0062 & 0.0187 \\
IQ $\uparrow$ & 20 & 0.7529 & 0.7507 & 0.7514 & 0.7604 & 0.7475 & 0.7529 & 0.0043 & 0.0129 & 0.0166 \\
AQ $\uparrow$ & 20 & 0.5354 & 0.5353 & 0.5390 & 0.5391 & 0.5319 & 0.5350 & 0.0027 & 0.0071 & 0.0243 \\
Cam $\uparrow$ & 20 & 83.55 & 83.16 & 82.85 & 81.89 & 81.84 & 81.76 & 0.78 & 1.79 & 11.01 \\
\bottomrule
\end{tabular}
\end{table}

Across the six seeds, mean Cam ranged from 81.76 to 83.55 out of 100, and mean MN-MS ranged from 0.9567 to 0.9629. Table~\ref{tab:mc_joint_inference_seeds} gives the range across seeds for each metric, computed both across the six means and within each input. The final column gives the median within-input range.

\end{document}